\documentclass[runningheads,a4paper]{llncs}
\usepackage[english]{babel}
\usepackage[utf8]{inputenc}
\usepackage{amsmath}
\usepackage{graphicx}
\usepackage{subcaption}
\usepackage[colorinlistoftodos]{todonotes}
\usepackage{algorithm}
\usepackage{algorithmic}
\usepackage{array}
\usepackage{wrapfig}
\usepackage{booktabs}

\usepackage{subcaption}
\title{Deep Residual Learning for Instrument Segmentation in Robotic Surgery}
\author{Daniil Pakhomov$^{1,3}$, Vittal Premachandran$^{1}$, Max Allan$^{2}$, \\ Mahdi Azizian$^{2}$ and Nassir Navab$^{1,3}$}
\institute{$^{1}$ Johns Hopkins University, USA \\
$^{2}$ Intuitive Surgical Inc., USA \\
$^{3}$ Technische Universität München, Germany}

\authorrunning{ }

\begin{document}
\maketitle

\begin{abstract}
Detection, tracking, and pose estimation of surgical instruments are crucial tasks for computer assistance during minimally invasive robotic surgery. In the majority of cases, the first step is the automatic segmentation of surgical tools. Prior work has focused on binary segmentation, where the objective is to label every pixel in an image as tool or background. We improve upon previous work in two major ways. First, we leverage recent techniques such as deep residual learning and dilated convolutions to advance binary-segmentation performance. Second, we extend the approach to multi-class segmentation, which lets us segment different parts of the tool, in addition to background. We demonstrate the performance of this method on the MICCAI Endoscopic Vision Challenge Robotic Instruments dataset. The source code for the experiments reported in the paper has been made public\footnote{https://github.com/warmspringwinds/tf-image-segmentation}.
\end{abstract}

\section{Introduction}

Robot-assisted Minimally Invasive Surgery (RMIS) overcomes many of the limitations of traditional laparoscopic Minimally Invasive Surgery (MIS), providing the surgeon with improved control over the anatomy with articulated instruments and dexterous master manipulators. In addition to this, 3D-HD visualization on systems such as da Vinci enhances the surgeon's depth perception and operating precision~\cite{bhayani2005three}. However, complications due to the reduced field-of-view provided by the surgical camera limit the surgeon's ability to self-localize. Traditional haptic cues on tissue composition are lost through the robotic control system~\cite{okamura2009haptic}.

Overlaying pre- and intra-operative imaging  with the surgical console can provide the surgeon with valuable information which can improve decision making during complex procedures~\cite{taylor2008medical}. However, integrating this data is a complex task and involves understanding spatial relationships between the surgical camera, operating instruments and patient anatomy. A critical component of this process is segmentation of the instruments in the camera images which can be used to prevent rendered overlays from occluding the instruments while providing crucial input to instrument tracking frameworks \cite{pezzementi2009articulated,allan2014d}.
\begin{figure}[!tbp]
  \centering
  \begin{minipage}[b]{0.49\textwidth}
  \centering
    \includegraphics[width=\textwidth, trim={5cm 0cm 5cm 0cm},clip, height=6.85cm]{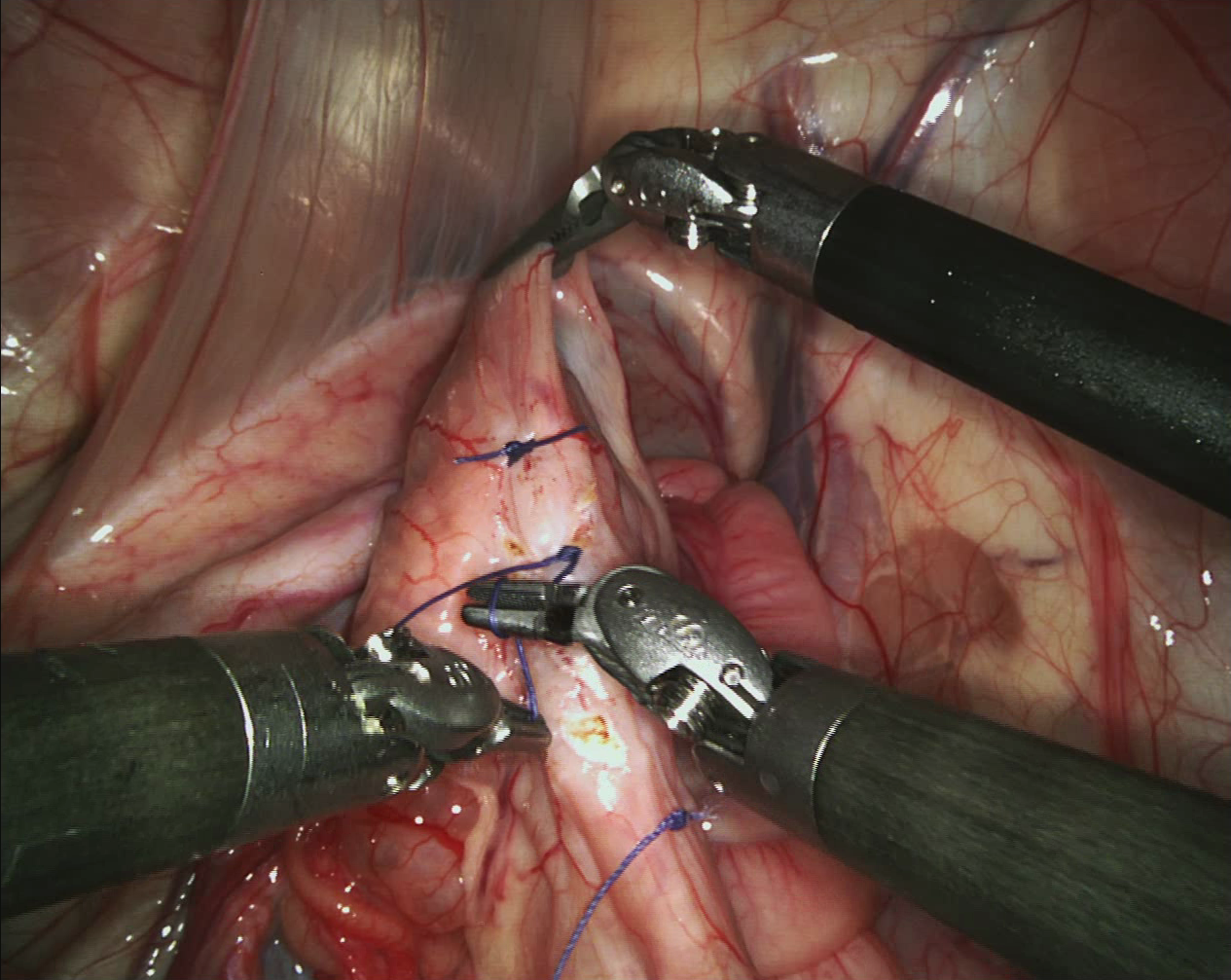}
    \caption{\label{fig:all_seg_frames} Example frames from RMIS procedures demonstrating the complex lighting and color distributions which make instrument segmentation an extremely challenging problem.}
  \end{minipage}
  \hfill
  \begin{minipage}[b]{0.49\textwidth}
  \centering
    \includegraphics[trim={12cm 6cm 13cm 0cm},clip, height=4cm]{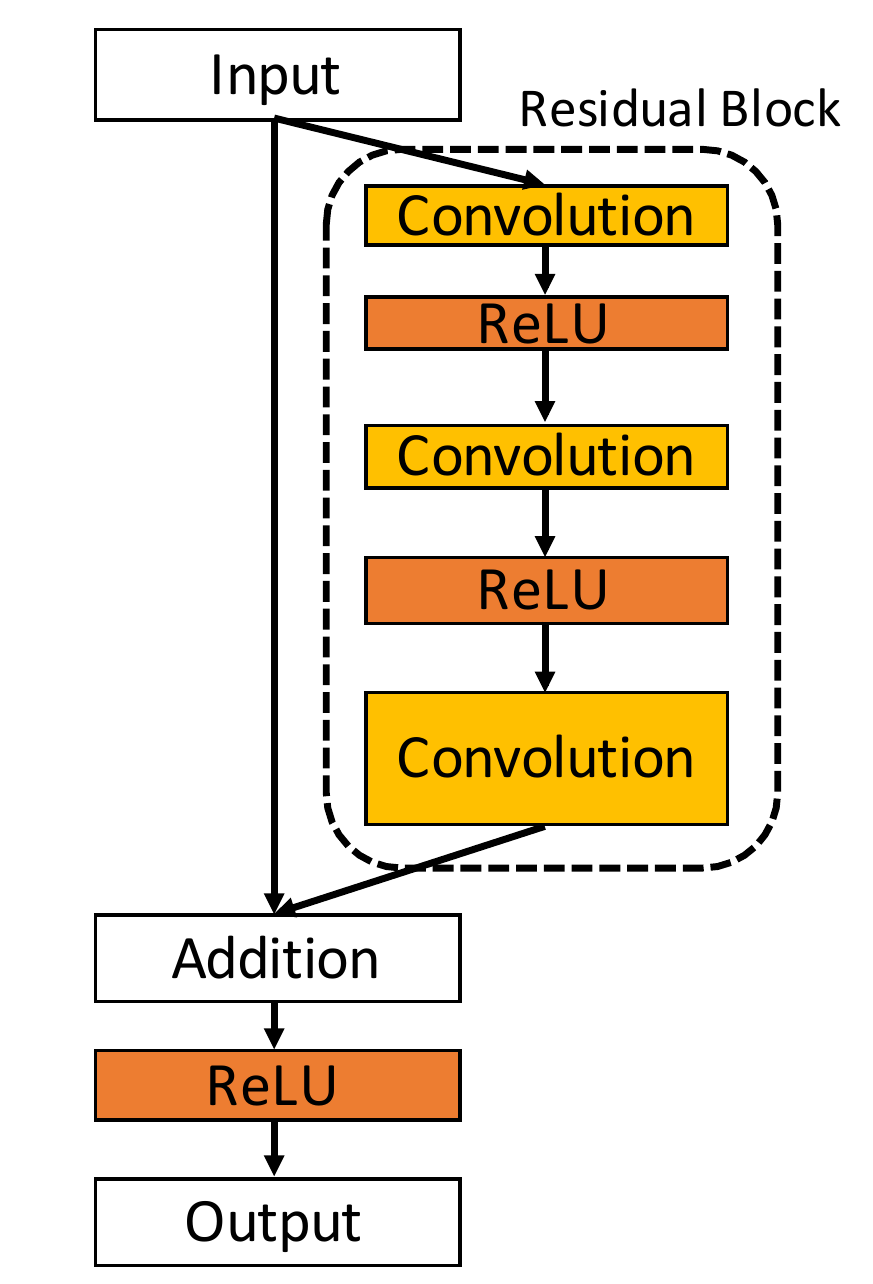}
    \caption{\label{fig:resnet_blk}Architecture of a "bottleneck" residual block which is composed of three convolution layers. The first convolutional layer performs dimensionality reduction, leaving the  middle layer with smaller
input/output dimensions and the third convolutional layer expands the dimension back to the original size. The output of the third convolutional layer is the residual, which is added to the input features. Batch normalization was omitted for simplicity.}
  \end{minipage}
  \vspace{-4ex}
\end{figure}






Segmentation of surgical tools from tissue backgrounds is an extremely difficult task due to lighting challenges such as shadows and specular reflections, visual occlusions such as smoke and blood, and due to complex background textures (see Fig.~\ref{fig:all_seg_frames}). Early methods attempted to simplify the problem by modifying the appearance of the instruments \cite{tonet2005tracking}. However, this complicates clinical application of the technique as sterilization can become an issue. Segmentation of the instruments using natural appearance is a more desirable approach as it can be applied directly to pre-existing clinical setups. However, this defines a more challenging problem. To solve it, previous work has relied on machine learning techniques to model the complex discriminative boundary. The instrument-background segmentation can be modeled as a binary segmentation problem to which discriminative models, such as Random Forests \cite{bouget2015detecting}, maximum likelihood Gaussian Mixture Models \cite{pezzementi2009articulated} and Naive Bayesian classifiers \cite{speidel2006tracking}, all trained on color features, have been applied. A more recent work, showing state-of-the-art performance~\cite{garciareal}, applies Fully Convolutional Networks (FCNs), more specifically FCN-8s model~\cite{long2015fully} for the task of binary segmentation of robotic tools. Although most approaches treat the problem as a binary segmentation problem, for different applications of instrument tracking, 
it is important to discriminate between different parts of the instrument, particularly the rigid shaft and the metallic clasper \cite{allan2014d}. To the best of our knowledge, no previous work has performed multi-class robotic tool segmentation on the MICCAI Endoscopic Vision Challenge Robotic Instruments dataset \cite{med_segm_dataset}.

In this work, we adopt the state-of-art  residual image
classification Convolutional Neural Network (CNN)~\cite{he2016deep} for the task of semantic image segmentation by casting it into Fully Convolutional Network (FCN)~\cite{long2015fully}.
However, the transformed model delivers prediction map of significantly reduced dimension compared to the input image~\cite{long2015fully}.
To account for that and recover full resolution feature map, we reduce in-network downsampling, employ
dilated (atrous) convolutions to enable initialization with the parameters of the original classification network, and perform simple bilinear interpolation
of the feature maps to obtain the original image size~\cite{chen2016deeplab}~\cite{yu2015multi}. This
approach is a powerful alternative to using
deconvolutional layers and ``skip architecture" as in FCN-8s model~\cite{long2015fully}. By employing it, we advance the state-of-the-art in binary segmentation of tools in the aforementioned dataset and extend our approach for multi-class tool segmentation.

\section{Method}
The goal of this work is to label every pixel of an image $\textbf{I}$ with one of $C$  semantic classes, representing surgical tool part or background. In case of binary segmentation, the goal is to label each pixel into $C=2$ classes, namely surgical tool and background. In this work, we also consider a more challenging multi-class segmentation with $C=3$ classes, namely tool's shaft, tool's manipulator and background.

Each image $\textbf{I}_i$ is a three-dimensional
array of size $h \times w \times d$, where $h$ and $w$ are spatial dimensions,
and $d$ is a channel dimension. 
In our case, $d=3$ because we use RGB images. Each image $\textbf{I}_i$ in the training dataset has corresponding annotation $\textbf{A}_i$ of a size $h \times w \times C$ where each element represents
one-hot encoded semantic label $a \in \{0, 1\}^C$ (for example, if we
have classes 1, 2, and 3, then the one-hot encoding of label 2 is $(0, 1, 0)^{T}$).

We aim at learning a mapping from $\textbf{I}$ to $\textbf{A}$ in a supervised fashion that generalizes to previously unseen images. In this work, we use CNNs to learn a discriminative classifier which delivers pixel-wise predictions given an input image. Our method is built upon state-of-the-art deep residual image classification CNN (ResNet-101, Section \ref{sec:residual}), which we convert into fully convolutional network (FCN, Section \ref{sec:FCN}).

CNNs reduce the spatial resolution of the feature maps by using pooling layers or convolutional layers with strides greater than one. However, for our task of pixel-wise prediction we would like dense feature maps.
We set the stride to one in the last two layers responsible for downsampling, and in order to reuse the weights from a pre-trained model, we dilate the subsequent convolutions (Sec. \ref{sec:dilated}) with an appropriate rate. This enables us to obtain predictions that are downsampled only by a factor of $8\times$ (in comparison to the original downsampling of $32\times$).

We then apply  bilinear interpolation to regain the original spatial resolution. 
With an output map of the same resolution as an input image, we perform end-to-end training by minimizing the normalized pixel-wise cross-entropy loss~\cite{long2015fully}.

\subsection{Deep Residual Learning}
\label{sec:residual}

Traditional convolutional networks learn filters that process the input $x_l$ and produce a filtered response $x_{l+1}$, as shown below
\begin{gather}
y_l = g(x_l, w_l),
\\
x_{l+1} = f(y_l).
\end{gather}
Here, $g(.,.)$ is a standard convolutional layer with $w_l$ being the weights of the layer's convolutional filters and biases, $f(.)$ is a non-linear mapping function such as the Rectified Linear Unit (ReLU). Many state of the art CNNs employ such convolutional layer followed by a non-linear rectification as a basic building block (AlexNet, VGG16, etc.). However, He et al. \cite{he2016deep} recently showed that significant gains in performance can be obtained by employing ``residual units" as a building block of a deep CNN, and called such networks Residual Networks (ResNets). In this work, we use a residual network to perform image segmentation. Deep residual networks (ResNets)~\cite{he2016deep} consist of many stacked “Residual Units”. Fig. \ref{fig:resnet_blk} shows the architecture of a residual unit. Each unit can be expressed in the following general form,
\begin{gather}
y_l = h(x_l) + F(x_l
, W_l),
\\
x_{l+1} = f(y_l),
\end{gather}
where $x_l$ and $x_{l+1}$ are input and output of the $l$-th unit, and $F(.,)$ is a residual function to be learnt. In~\cite{he2016deep}, the function $h(.)$ is a simple identity mapping, $h(x_l) = x_l$ and $f(.)$ is a rectified linear unit activation (ReLU) function. Because $h(x_l)$ is chosen to be an identity mapping, it is easily realized by attaching an identity skip connection (also known as a ``shortcut” connection).

Assuming that the desired
underlying mapping for $y_l$ is $H(x_l)$, a residual block fits a mapping of $F(x_l) = H(x_l)-x_l$, which is called a residual function. The original
mapping is recast into $F(x_l)+x_l$.
It was experimentally shown in \cite{he2016deep} that learning residual functions with reference to the layer inputs, instead
of learning unreferenced functions allows to train deeper models which gain accuracy from considerably increased depth.


ResNets that are over 100-layers deep have shown to produce state-of-the-art accuracy for several challenging Image Classification and Image Segmentation tasks~\cite{he2016deep}~\cite{chen2016deeplab}. This motivates our choice of using ResNet architecture over others. In our work, we adopt ResNet-101 architecture for Image Segmentation and apply it for the task of tool segmentation.

\subsection{Fully Convolutional Networks} \label{sec:FCN}

\begin{figure*}[!htb]
    \centering
		\includegraphics[width=1\textwidth]{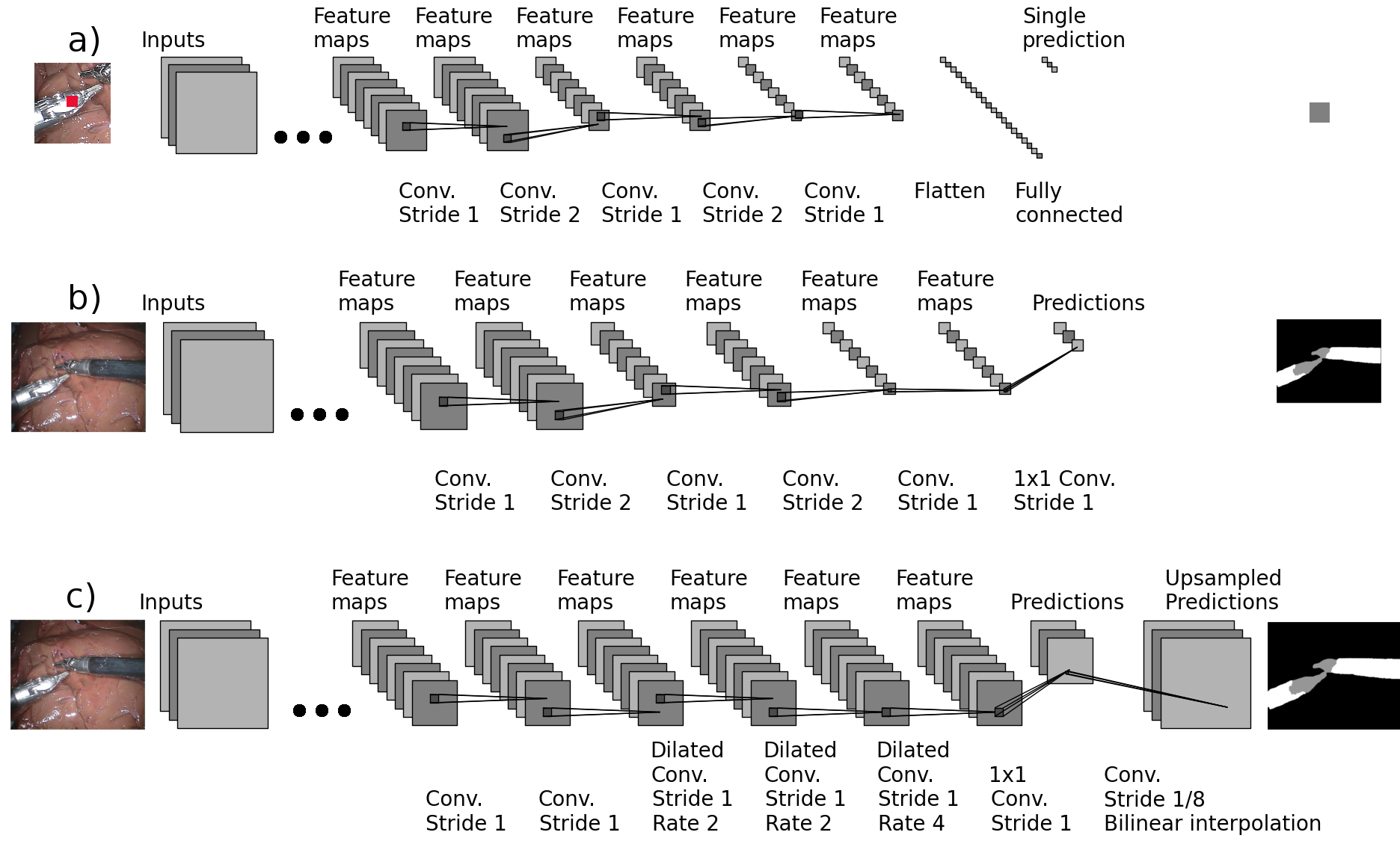}
\caption{\label{fig:fcn_and_dilated}A simplified CNN before and after being converted into an FCN (illustrations \textbf{(a)} and \textbf{(b)} respectively), after reducing downsampling rate with integration of dilated convolutions into its architecture with subsequent bilinear interpolation (illustration \textbf{(c)}). Illustration \textbf{(a)} shows an example of applying a CNN to an image patch centered at the red pixel which gives a single vector of predicted class scores (manipulator, shaft, and background). Illustration \textbf{(b)} shows the fully connected layer being converted into $1 \times 1$ convolutional layer, making the network fully convolutional, thus enabling a dense prediction. Illustration \textbf{(c)} shows network with reduced downsampling and dilated convolutions that produces outputs that are being upsampled to acquire pixelwise predictions.}
\vspace{-5ex}
\end{figure*}

Deep CNNs (e.g. AlexNet, VGG16, ResNets, etc.) are primarily trained for the task of image classification. Hence, they are originally designed to solve recognition problems on the scale of entire image, by assigning one of many class labels to it. However, to obtain the output granularity required for a task such as image segmentation the network should be modified. This modification consists of converting fully connected layers into convolutions with kernels that are equal to their fixed input regions~\cite{long2015fully}. Such a network is called a Fully Convolutional Network (FCN). FCN operates on inputs of any size, and produces an output with reduced spatial dimensions~\cite{long2015fully}. The reduction in the spatial dimension is due to the presence of either pooling (VGG16) or convolutional (ResNets) layers with a stride greater than one pixel.

In order to convert our Image Classification CNN (ResNet-101) into FCN we follow the recent line of work by Long et al.~\cite{long2015fully} and Chen et al.~\cite{chen2016deeplab} by removing the final average pooling layer and replacing the fully connected layer with a $1 \times 1$ convolutional layer.
Doing so casts the network into FCN that takes input of any size and produces an output with predictions over a spatial grid of smaller resolution. This transformation is illustrated in Fig.~\ref{fig:fcn_and_dilated}b.

Fully convolutional models deliver prediction maps with significantly reduced dimensions (for both VGG16 and ResNets, the spatial dimensions are reduced by a factor of $32$). In the previous work~\cite{long2015fully}, it was shown that adding a deconvolutional layer to learn the upsampling with factor $32$ provides a way to get the prediction map of original image dimension, but the segmentation boundaries delivered by this approach are usually too coarse. To tackle this problem, two approaches were recently developed which are based on modifying the architecture. (i) By fusing features from layers of different resolution to make the predictions~\cite{long2015fully}. (ii) By avoiding downsampling of some of the feature maps~\cite{chen2016deeplab}~\cite{yu2015multi} (removing certain pooling layers in VGG16 and by setting the strides to one in certain convolutional layers responsible for the downsampling in ResNets). However, since the weights in the subsequent layers were trained to work on a downsampled feature map, they need to be adapted to work on the feature maps of a higher spatial resolution. To this end,~\cite{chen2016deeplab} employs dilated convolutions. In our work, we follow the second approach: we mitigate the decrease in the spatial resolution by using convolutions with strides equal to one in the last two convolutional layers responsible for downsampling in ResNet-101 and by employing dilated convolutions for subsequent convolutional layers (Sec. \ref{sec:dilated}).

\subsection{Dilated Convolutions}
\label{sec:dilated}

In order to account for the problem stated in the previous section, we use dilated (atrous) convolution. Dilated convolution\footnote{We follow the practice of previous work and use
simplified definition without mirroring and centering the filter~\cite{chen2016deeplab}.} in one-dimensional case is defined as 
$$
y[i] = \sum_{k=1}^{K}x[i + rk]w[k]
$$
where, $x$ is an input 1D signal, $y$ output signal and $w$ is a filter of size $K$.
The rate parameter $r$ corresponds to the dilation factor. The dilated convolution operator can reuse the weights from the filters that were trained on downsampled feature maps by sampling the unreduced feature maps with an appropriate rate.

In our work, since we choose not to downsample in some convolutional layers (by setting their stride to one instead of two), convolutions in
all subsequent layers are dilated. This enables initialization with the parameters of the original classification network, while producing higher-resolution outputs. This transformation follows~\cite{chen2016deeplab} and is illustrated in Fig.~\ref{fig:fcn_and_dilated}c.

\subsection{Training} \label{Training}

Given a sequence of images $\{ \textbf{I}_t \}^{n_t}_{t=0}$, and sequence of ground-truth segmentation annotations 
$\{ \textbf{A}_t \}^{n_t}_{t=0}$, we optimize normalized pixel-wise cross-entropy loss~\cite{long2015fully} using Adam optimization algorithm~\cite{kingma2014adam} with learning rate set to $10^{-4}$ ($n_t$ stands for the number of training examples). We choose the learning rate of $10^{-4}$ after performing a grid search over five different learning rates and found that $10^{-4}$ helps produce the best score on the validation dataset. Other parameters of Adam optimization algorithm were set to the values suggested in \cite{kingma2014adam}.



\section{Experiments and Results}

\begin{table}
\centering
\begin{tabular}{c|c|c|c}
			\toprule
	        & Sensitivity & Specificity & \shortstack{Balanced \\ Accuracy}\\ \hline
			FCN-8s \cite{garciareal} & $87.8 \%$  & $88.7 \%$  & $88.3 \%$  \\
			Our work & $85.7 \%$ & $98.8 \% $ & $\textbf{92.3\%}$        
  		\end{tabular}
    \caption{\label{tab:comparison_table}Shows comparison of our results with previous state-of-the art~\cite{garciareal} in binary segmentation of robotic tools. Our method provides a 4\% improvement in balanced accuracy.}
\end{table} 

We test our method on the MICCAI Endoscopic Vision Challenge's Robotic Instruments dataset. This dataset consists of four $45$-second 2D stereo image sequences with Large Needle Driver (LND) instruments in an ex-vivo setup that is used for training. Each pixel is labeled as either background, shaft or articulated head. The test data consists of four $15$-second sequences with similar background to training sequence.
Two $1$-minute 2D image sequences of $2$ instruments in an ex-vivo setup are also in the test dataset. These sequences also contain tool that is not present in the training dataset. The sequences contain occlusions and articulations.
\vspace{-4ex}
\begin{figure}
\centering     
\begin{subfigure}{0.24\linewidth}%
        \includegraphics[width=\linewidth]{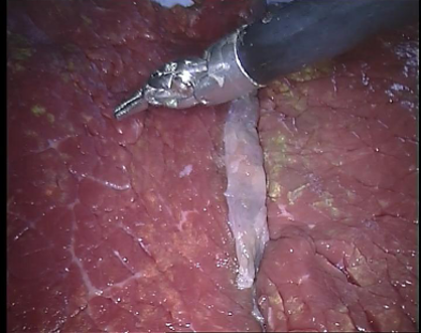}%
        \vspace{0.9ex}
        \caption{}%
\end{subfigure}%
\begin{subfigure}{0.24\linewidth}
        \includegraphics[width=\linewidth]{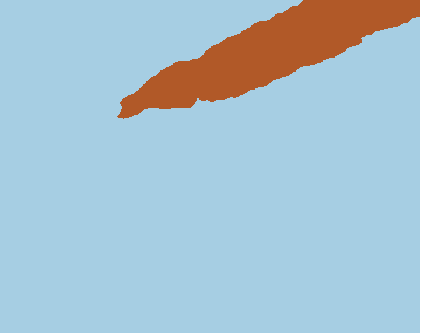}
        \vspace{-2ex}
        \caption{}
\end{subfigure}%
\begin{subfigure}{0.24\linewidth}
        \includegraphics[width=\linewidth]{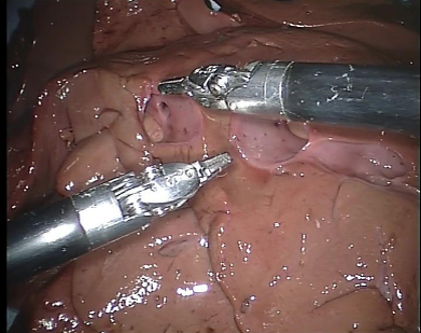}
        \vspace{-2ex}
        \caption{}
\end{subfigure}%
\begin{subfigure}{0.24\linewidth}
        \includegraphics[width=\linewidth]{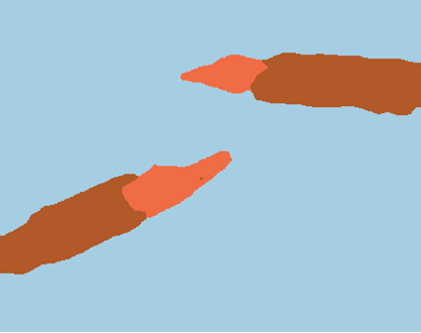}
        \vspace{-2ex}
        \caption{}
\end{subfigure}
\vspace{-2ex}
\caption{Figure shows some qualitative results from our method. (a) and (c) show example image frames from the dataset. (b) shows the binary segmentation output of the image in (a). (d) shows the multi-class segmentation output of the image in (c).}
\label{fig:qual}
\vspace{-5ex}
\end{figure}


\subsection{Results}

\begin{table}
\centering
		\begin{tabular}{c|c|c|c|c}
        \toprule
	        &  C1 & C2 & C3 & \shortstack{Overall\\Mean}\\ \hline
			Video 1 & 79.6 & 68.2  & 96.5 & 81.4  \\
			Video 2 & 82.2 & 70.2 & 98.6 & 83.7 \\
			Video 3 & 80.4 & 66.4 & 98.0 & 81.6 \\
			Video 4 & 75.0 & 44.9 & 97.1 & 72.3 \\
			Video 5 & 72.3 & 56.0  & 96.3 & 74.9  \\
			Video 6 & 70.7 & 50.2  & 95.7 & 72.2       
  		\end{tabular}
    \caption{\label{tab:iou} Table reports IoU results for the task of multi-class image segmentation. C1, C2 and C3 corresponds to  Manipulator, Shaft, and Background, respectively.}
\end{table}

We report our results in Tab. \ref{tab:comparison_table} using standard metrics such as sensitivity and specificity and compare with the previous state-of-the-art \cite{garciareal} for the task of binary segmentation. We can see that our method outperforms the previous work by 4\%. We also report results using the Intersection Over Union (IoU) metric for the task of multi-class segmentation in Tab. \ref{tab:iou}. IoU is a standard metric used for quantifying segmentation results \cite{everingham2010pascal}. To the best of our knowledge, we are the first to report segmentation results on the multi-class segmentation task on this dataset. Fig. \ref{fig:qual} shows some qualitative results for both the binary segmentation and the multi-class segmentation tasks.

\section{Discussion and Conclusion}

In this work, we propose a method to perform robotic tool segmentation. This is an important task, as it can be used to prevent rendered overlays from occluding the instruments or to estimate the pose of a tool~\cite{allan2014d}. We use deep network to model the mapping from the raw images to the segmentation maps. Our use of a state-of-the-art deep network (ResNet-101) with dilated convolutions helps us achieve 4\% improvement in binary tool segmentation over the previous stat-of-the-art. In addition, we extend the binary segmentation task to multi-class segmentation task (segmenting out tool parts). We are the first to do this on the MICCAI Endoscopic Vision Challenge's Robotic Instruments dataset.  Our results show the benefit of using deep residual networks for this task and also provide a solid baseline for future work on multi-class segmentation.



\bibliographystyle{plain}
\bibliography{bibliography.bib}

\end{document}